\documentclass{article}

\usepackage{microtype}
\usepackage{graphicx}
\usepackage{subcaption}
\usepackage{booktabs}
\usepackage{hyperref}
\usepackage{amsmath, amssymb, mathtools}
\usepackage{xcolor}
\usepackage[capitalize,noabbrev]{cleveref}
\usepackage{multirow}
\usepackage{natbib}
\usepackage{threeparttable}

\usepackage[preprint]{icml2026}

\usepackage{xurl}
\newcommand{\hf}[1]{\href{https://huggingface.co/#1}{\nolinkurl{#1}}}

\icmltitlerunning{The Viscosity of Logic}

\newcommand{\Mistral}{Mistral-7B}
\newcommand{\Llama}{LLaMA-2-7B}
\newcommand{\Qwen}{Qwen-1.5-7B}

\begin{document}

\twocolumn[
\icmltitle{The Viscosity of Logic: Phase Transitions and Hysteresis in DPO Alignment}

\begin{icmlauthorlist}
\icmlauthor{Marco Pollanen}{inst1}
\icmlaffiliation{inst1}{Department of Mathematics, Trent University, Peterborough, ON, Canada}
\icmlcorrespondingauthor{Marco Pollanen}{marcopollanen@trentu.ca}
\end{icmlauthorlist}

\icmlkeywords{Alignment, DPO, Phase Transitions, Hysteresis, Large Language Models}

\vskip 0.3in
]

\printAffiliationsAndNotice{}

\begin{abstract}
Direct Preference Optimization (DPO) is often tuned as if increasing alignment pressure (controlled by $\beta$) yields progressively ``better'' behavior.
We instead treat $\beta$ as a control parameter and densely sweep it for three 7B open-weight families under a fixed DPO recipe.
In \Mistral{}, capability is sharply non-monotonic: aggregated logic-probe margins become positive only in a narrow band near $\beta \approx 10^{-2}$ and revert outside it, with boundary points that are seed-sensitive.
Across architectures under the same sweep, we observe qualitatively different response modes: sharp reorganization in \Mistral{}, selective changes in \Llama{}, and smooth trade-offs in \Qwen{}.
Critically, the DPO preference margin can anticorrelate with reasoning capability (Pearson $r=-0.91$ for \Llama{} logic), so margin-based selection can prefer capability-impaired models.
Training path also matters: exposure to high $\beta$ induces capability losses that persist even after $\beta$ is reduced (hysteresis).
These findings motivate capability-resolved evaluation across the $\beta$ landscape rather than reliance on margins or aggregate benchmarks.
\end{abstract}

%%%%%%%%%%%%%%%%%%%%%%%%%%%%%%%%%%%%%%%%%%%%%%%%%%%%%%%%%%%%%%%%%%%%%%%%%%%%%%%
\section{Introduction}
%%%%%%%%%%%%%%%%%%%%%%%%%%%%%%%%%%%%%%%%%%%%%%%%%%%%%%%%%%%%%%%%%%%%%%%%%%%%%%%

\begin{figure}[!t]
\centering
\includegraphics[width=\columnwidth]{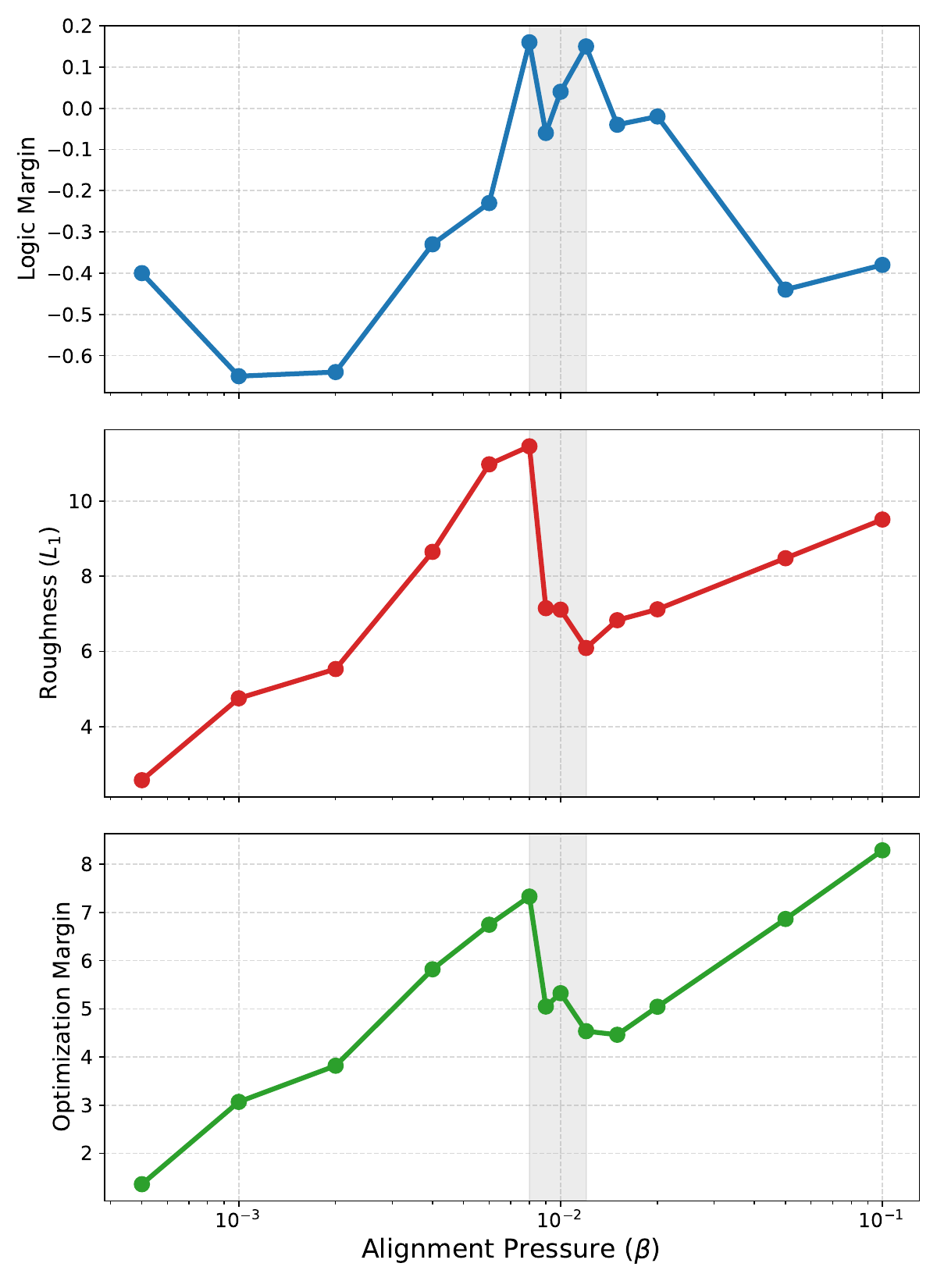}
\caption{
\textbf{Alignment is not monotonic---it has sharp transitions.}
Sweeping alignment pressure $\beta$ in \Mistral{} reveals three co-localized anomalies within a narrow band (shaded).
\textbf{Top:} Logic-probe margin is negative at low and high $\beta$, turning positive only at discrete points within a narrow band near $\beta \approx 10^{-2}$ (e.g., $\beta \in \{0.008, 0.010, 0.012\}$ in the canonical run). Boundaries are seed-sensitive: identical $\beta$ can yield opposite signs across runs.
\textbf{Middle:} Training roughness peaks sharply then drops by 37\% in a single $\beta$ step (0.008$\rightarrow$0.009), consistent with a sudden change in optimization dynamics.
\textbf{Bottom:} The DPO preference margin \emph{decreases} inside the logic-positive pocket, so margin-based selection would reject the most logic-positive (in the canonical run) regime.
The alignment landscape has cliffs, not just slopes.
}
\label{fig:grand_unified}
\end{figure}

The comforting narrative---stronger alignment, better behavior---is dangerously underspecified. Our goal is not to propose a tuning recipe, but to map a failure mode: at fixed model scale, alignment pressure can induce abrupt capability reorganization.
We treat $\beta$ as a control parameter and empirically chart the resulting capability landscape. Concretely, we run a controlled DPO sweep where each $\beta$ point starts from the same base checkpoint (except in explicit hysteresis tests).
We evaluate both the optimized DPO preference margin and capability-specific probe margins, and we repeat evaluations across random seeds near sensitive regions.

In standard practice, one sweeps $\beta$, selects the checkpoint with the best margin, and deploys. This pipeline can fail in at least three ways:
\begin{enumerate}
\item Proxy failure: the DPO preference margin can \emph{anticorrelate} with capability (e.g., $r=-0.91$ for \Llama{} logic).
\item Local instability: near sensitive $\beta$, neighboring grid points and different random seeds can yield qualitatively different outcomes at identical hyperparameters.
\item Path dependence: exposure to higher $\beta$ can induce persistent capability degradation even after $\beta$ is reduced (hysteresis).
\end{enumerate}

These are empirical observations from controlled experiments across three 7B architectures, with \Mistral{} exhibiting a sharp functional transition and \Llama{} and \Qwen{} providing contrasting selective and smooth responses.

\paragraph{Organizing lens.}
We borrow ``phase transition'' and ``hysteresis'' as operational descriptors for (i) sharp qualitative changes over a narrow $\beta$ interval, (ii) elevated seed sensitivity near a boundary, and (iii) path-dependent final capability at identical terminal $\beta$.
We use ``viscosity'' informally to denote architecture- and capability-specific resistance to such reorganization under our fixed training recipe.

\paragraph{Terminology.}
We use \emph{logic-positive pocket} to denote the narrow $\beta$ band in which the aggregated logic \emph{probe} margin is positive in at least one run under our protocol (e.g., the canonical seed), while allowing that the sign can flip across seeds near (and sometimes within) this band.
Accordingly, we treat the pocket as \emph{seed-sensitive} rather than uniformly ``ordered.''

\paragraph{Contributions.}
We document these phenomena through canonical phase diagrams (\S\ref{sec:results:emergence}), architecture comparisons (\S\ref{sec:results:architecture}), margin-capability analysis (\S\ref{sec:results:decoupling}), hysteresis tests (\S\ref{sec:results:hysteresis}), probe decomposition (\S\ref{sec:results:probes}), and evidence for a shared sensitivity region across architectures (\S\ref{sec:results:sensitivity}).

Reading guide: Figure~\ref{fig:grand_unified} shows the core non-monotonicity in \Mistral{}; \Cref{sec:results:decoupling} establishes proxy--capability anticorrelation; \Cref{sec:results:hysteresis} demonstrates path dependence.

%%%%%%%%%%%%%%%%%%%%%%%%%%%%%%%%%%%%%%%%%%%%%%%%%%%%%%%%%%%%%%%%%%%%%%%%%%%%%%%
\section{Background and Setup}
%%%%%%%%%%%%%%%%%%%%%%%%%%%%%%%%%%%%%%%%%%%%%%%%%%%%%%%%%%%%%%%%%%%%%%%%%%%%%%%

\subsection{Direct Preference Optimization}

DPO \citep{rafailov2023direct} optimizes pairwise preferences with a Kullback--Leibler (KL) divergence term relative to a reference policy and a temperature-like scalar $\beta$.
In the standard DPO objective, $\beta$ scales the preference term; in common derivations and implementations it is inversely related to the effective strength of staying close to the reference policy.
Accordingly, we treat larger $\beta$ as stronger preference-optimization pressure and refer to it as ``higher alignment pressure'' in this paper.
We use two distinct margin notions throughout: 
(i) the \emph{DPO preference margin} (the optimized proxy), which we denote by $m_t$ over training steps and aggregate at the end of training;
and (ii) \emph{probe margins}, length-normalized log-probability differences on fixed capability probes, where positive values indicate correct behavior under the probe scoring rule.
Standard practice tunes $\beta$ for downstream performance \citep{tunstall2023zephyr}; we instead sweep it systematically to map the capability landscape.

\paragraph{Ruling out metric artifacts.}
Apparent emergence can arise from nonlinear evaluation metrics \citep{schaeffer2023emergent}.
Our key discontinuity evidence is \emph{path dependence}: two training paths that end at identical $\beta$ are evaluated identically yet yield measurably different final probe margins, implying different final parameter states under our protocol.
While DPO variants can exhibit length effects \citep{park2024disentangling,xu2024contrastive}, our main qualitative results concern sign changes and path dependence in fixed probes rather than aggregate generation length.

\subsection{Experimental Protocol}

\paragraph{Frozen configuration principle.}
Each $\beta$ point begins from a fresh base model copy with identical recipe, isolating $\beta$ effects from confounds.
Deviations (e.g., hysteresis tests) are deliberate and labeled.

\paragraph{Models.}
We selected three 7B families to span the open-weight ecosystem rather than to cherry-pick favorable results:
\Mistral{} (\texttt{mistralai/Mistral-7B-v0.1}),
\Llama{} (\texttt{NousResearch/Llama-2-7b-hf}), and
\Qwen{} (\texttt{Qwen/Qwen1.5-7B}).
Unless otherwise noted, we used identical hyperparameters across models (learning rate $5 \times 10^{-5}$, 200 steps, batch size 4, LoRA rank 8 with alpha 16 and dropout 0.05).

\paragraph{Beta selection.}
We employed a logarithmic sweep
\[
\beta \in 
\left\{
\begin{aligned}
&0.0005, 0.001, 0.002, 0.004, 0.006, 0.008, 0.009,\\
&0.010, 0.012, 0.015, 0.020, 0.050, 0.100
\end{aligned}
\right\},
\]
with increased resolution in the high-gradient region ($\beta \approx 0.006$--$0.015$), where preliminary coarse sweeps indicated heightened sensitivity (sharp roughness changes and elevated seed variance).

\paragraph{Capability probes.}
\textbf{Logic}: syllogistic reasoning, ordering (3 probes).
\textbf{Arithmetic}: multi-digit operations (3 probes).
\textbf{Format}: JSON generation, boolean constraints (4 probes).
\textbf{Sycophancy}: resistance to false claims (2 probes), following the evaluation framework of \citet{perez2022discovering}.
\textbf{Negation}: negated query handling (2 probes).
We compute length-normalized log-probability margins; positive margins indicate correct behavior.

\paragraph{Training roughness.}
We quantify optimization turbulence by \emph{roughness}, defined as the mean absolute
stepwise change in the preference-margin trajectory over training:
$R \;=\; \frac{1}{T-1}\sum_{t=2}^{T} \left| m_t - m_{t-1} \right|$,
where $m_t$ is the preference margin at optimizer step $t$ and $T$ is the number of steps.

%%%%%%%%%%%%%%%%%%%%%%%%%%%%%%%%%%%%%%%%%%%%%%%%%%%%%%%%%%%%%%%%%%%%%%%%%%%%%%%
\section{Results}
\label{sec:results}
%%%%%%%%%%%%%%%%%%%%%%%%%%%%%%%%%%%%%%%%%%%%%%%%%%%%%%%%%%%%%%%%%%%%%%%%%%%%%%%

\subsection{R1: Non-Monotonic Capability Dynamics}
\label{sec:results:emergence}

At fixed 7B scale, varying $\beta$ induces sharp, non-monotonic capability changes.
Table~\ref{tab:canonical} presents the canonical \Mistral{} sweep.

\begin{table}[t]
\caption{\Mistral{} canonical sweep. Bold: logic-positive pocket where the aggregated logic \emph{probe} margin is positive in the canonical run.
Neighboring points ($\beta=0.009$, $0.015$) are seed-sensitive and can flip sign.
\textbf{Margin} denotes the final DPO preference margin (optimized proxy), not a probe score.}
\label{tab:canonical}
\centering
\scriptsize
\setlength{\tabcolsep}{2.5pt}
\begin{tabular}{l|ccccc|c}
\toprule
$\beta$ & Logic & Arith & Format & Syco & Neg & Margin \\
\midrule
0.0005 & $-$0.40 & +2.81 & +4.44 & +1.09 & +4.25 & 1.4 \\
0.001 & $-$0.65 & +2.34 & $-$0.62 & $-$1.81 & +3.59 & 3.1 \\
0.002 & $-$0.64 & +2.75 & $-$0.13 & +1.56 & +5.12 & 3.8 \\
0.004 & $-$0.33 & +2.68 & $-$0.52 & +0.82 & +3.88 & 5.8 \\
0.006 & $-$0.23 & +2.66 & $-$1.11 & $-$0.70 & +3.75 & 6.7 \\
\textbf{0.008} & \textbf{+0.16} & +2.67 & $-$1.11 & +1.29 & +3.61 & 7.3 \\
0.009 & $-$0.06 & +2.71 & $-$1.00 & +1.55 & +3.83 & 5.0 \\
\textbf{0.010} & \textbf{+0.04} & +2.75 & $-$0.90 & +1.25 & +3.66 & 5.3 \\
\textbf{0.012} & \textbf{+0.15} & +2.72 & $-$1.09 & +0.87 & +3.67 & 4.5 \\
0.015 & $-$0.04 & +2.68 & $-$1.14 & +1.30 & +3.48 & 4.5 \\
0.020 & $-$0.02 & +2.67 & $-$1.17 & +0.39 & +3.17 & 5.0 \\
0.050 & $-$0.44 & +2.65 & $-$1.24 & +0.40 & +2.47 & 6.9 \\
0.100 & $-$0.38 & +2.47 & $-$1.24 & $-$0.09 & +2.02 & 8.3 \\
\bottomrule
\end{tabular}
\end{table}

The key finding is a narrow \textbf{logic-positive pocket} in which the aggregated logic \emph{probe} margin becomes positive (in the canonical run, at $\beta \in \{0.008, 0.010, 0.012\}$).
The boundaries are seed-sensitive: at fixed $\beta$, different seeds can yield opposite signs, and neighboring grid points (e.g., $\beta = 0.009$, $0.015$) sit at the transition edge.
Multi-seed analysis reveals that variance peaks near $\beta = 0.006$, while the largest roughness discontinuity occurs at $\beta=0.008\rightarrow0.009$; together these signals localize a narrow critical region spanning roughly $\beta\approx0.006$--$0.009$ that coarse sweeps can easily miss.
Outside the pocket, aggregated logic-probe margins remain negative under our protocol.

\paragraph{Roughness collapse marks the boundary.}
The phase boundary is sharply marked by training dynamics: roughness peaks at $\beta = 0.008$ then collapses by 37\% at $\beta = 0.009$, the largest observed single-step transition in the sweep, computed directly from the logged roughness values at $\beta=0.008$ and $\beta=0.009$.
This roughness signature may be a useful \emph{screening signal} for boundary localization during hyperparameter search, to be confirmed by targeted multi-seed capability evaluation.

\paragraph{Hierarchy of response.}
Different capabilities respond at different pressures.
Format, sycophancy, and negation show positive margins at the lowest $\beta$ tested ($0.0005$); logic requires $\beta \approx 0.008$, an order of magnitude higher, suggesting a difficulty hierarchy where surface compliance is more readily achieved than deep reasoning.

\subsection{R2: Architecture-Dependent Responses}
\label{sec:results:architecture}

The same protocol yields qualitatively different regimes across architectures (Table~\ref{tab:three_arch}).

\begin{table}[t]
\caption{Three architectures at $\beta=0.01$: plastic, selective, smooth.}
\label{tab:three_arch}
\centering
\small
\begin{tabular}{lccc}
\toprule
& Mistral & LLaMA & Qwen \\
\midrule
Logic & +0.04 & $-$0.95 & +1.08 \\
Arith & +2.75 & +1.62 & +3.00 \\
Format & $-$0.90 & $-$1.98 & $-$0.70 \\
Syco & +1.25 & +1.71 & +3.01 \\
Neg & +3.66 & +1.58 & +0.42 \\
\midrule
Response & Plastic & Selective & Smooth \\
\bottomrule
\end{tabular}
\end{table}
We use these labels descriptively: \emph{plastic} denotes sharp qualitative reorganization with high seed sensitivity, \emph{selective} denotes capability-specific rigidity, and \emph{smooth} denotes continuous trade-offs without sharp discontinuities under our probe set.

To quantify sensitivity near the boundary, we measure seed variance at $\beta = 0.006$ across 5 seeds (Table~\ref{tab:variance}).

\begin{table}[t]
\caption{Variance at $\beta=0.006$ (5 seeds). \Mistral{} shows 27--2259$\times$ higher variance (Mistral/LLaMA) across the listed capabilities.}
\label{tab:variance}
\centering
\small
\setlength{\tabcolsep}{4pt} % default is 6pt
\begin{tabular}{lrrr}
\toprule
Capability & Mistral $\sigma^2$ & LLaMA $\sigma^2$ & Ratio (Mistral/LLaMA) \\
\midrule
Logic  & 0.016 & 0.0006  & 27$\times$   \\
Format & 0.015 & 0.00001 & 1751$\times$ \\
Syco   & 0.783 & 0.0003  & 2259$\times$ \\
\bottomrule
\end{tabular}
\end{table}

\begin{figure}[t]
\centering
\includegraphics[width=\columnwidth]{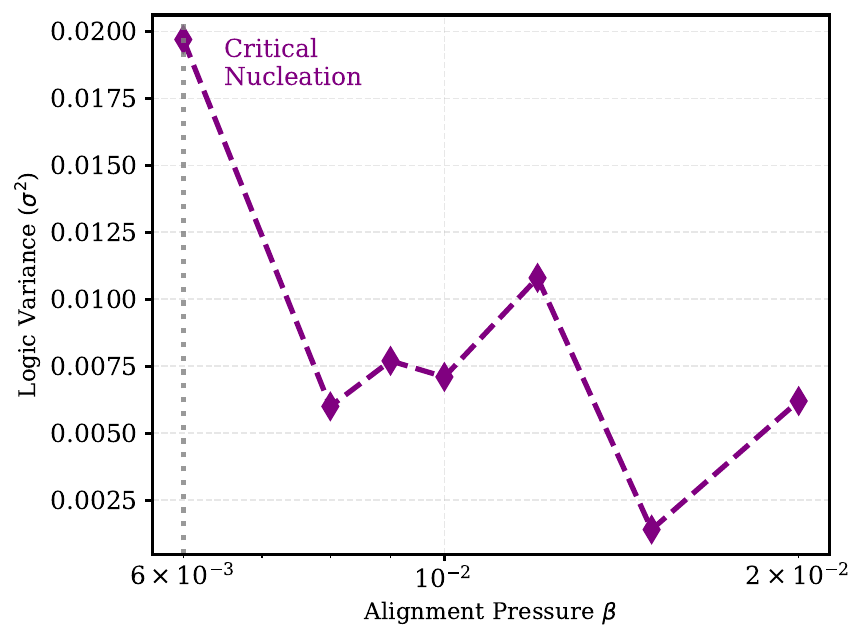}
\caption{\textbf{Critical fluctuations.} Logic variance across 5 seeds peaks at $\beta = 0.006$, marking the boundary region where outcomes are maximally seed-sensitive.}
\label{fig:variance}
\end{figure}

\Mistral{} exhibits 27--2259$\times$ higher variance than \Llama{} across logic, format, and sycophancy probes at $\beta=0.006$ (Figure~\ref{fig:variance}, Table~\ref{tab:variance}).
However, \Llama{}'s rigidity is capability-specific: while its logic remains frozen, sycophancy and format margins still vary with $\beta$, indicating compartmentalized response modes.

\paragraph{Seed-sensitive nucleation.}
High variance near the boundary means identical hyperparameters can yield opposite capability signs; single-run evaluations in this region are unreliable.

\paragraph{Learning-rate stress test (LLaMA).}
To test whether \Llama{}'s flat logic profile reflects insufficient step size, we varied the learning rate by 20$\times$ while holding all other settings fixed: $1\times 10^{-5}$, $5\times 10^{-5}$ (canonical), and $2\times 10^{-4}$.
We evaluated logic at $\beta \in \{0.006, 0.01, 0.02\}$ for each learning rate.
In the logged stress-test runs (seed 1), the aggregated logic margin remained negative for all three $\beta$ values at all three learning rates, suggesting that increased step size does not ``melt'' logic under this protocol.

\subsection{R3: Margin-Capability Decoupling}
\label{sec:results:decoupling}

A key practical risk is that the optimized DPO proxy can decouple from capability.
Models can appear more aligned by standard metrics while becoming less capable on reasoning tasks.

\begin{table}[t]
\caption{Margin-capability correlation. For \Llama{} logic: $r = -0.91$.}
\label{tab:correlation}
\centering
\small
\begin{threeparttable}
\begin{tabular}{lcc}
\toprule
Capability & Mistral & LLaMA \\
\midrule
Logic  & +0.27 & $-$0.91 \\
Format & $-$0.70 & --- \\
Neg    & $-$0.64 & --- \\
\bottomrule
\end{tabular}
\begin{tablenotes}[flushleft]
\scriptsize
\item \textbf{Note:} --- indicates correlation not reported because the corresponding sweep was not run or variance was insufficient to support a meaningful estimate under our protocol.
\end{tablenotes}
\end{threeparttable}
\end{table}

For \Llama{} logic, Pearson $r = -0.91$ (two-sided $p < 10^{-4}$, $n=13$), indicating that higher DPO preference margin is associated with lower logic-probe capability under our sweep.
This is consistent with an internal instance of Goodhart's law: increasing alignment pressure can improve the optimized proxy (margin) while degrading the measured target capability (logic).
This connects to reward overoptimization \citep{gao2023scaling}, but manifests here as capability destruction rather than reward hacking.

\begin{figure}[t]
\centering
\includegraphics[width=\columnwidth]{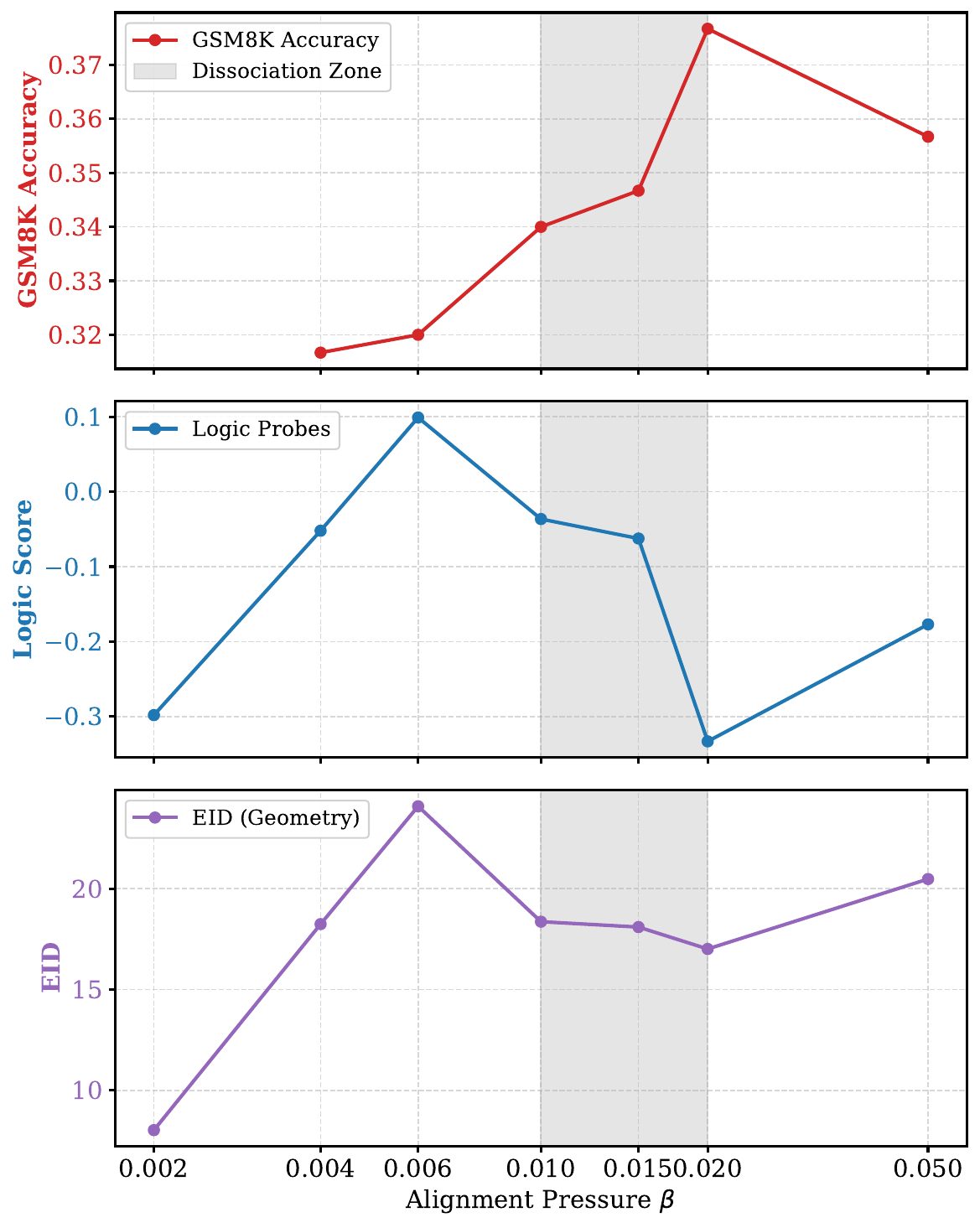}
\caption{\textbf{Capability--benchmark dissociation.} GSM8K peaks at $\beta = 0.02$ while logic probes decline. Shaded region: benchmarks improve despite internal degradation. This is Goodhart's law made visible.}
\label{fig:dissociation}
\end{figure}

Benchmark gains can coincide with internal probe degradation (Figure~\ref{fig:dissociation}).
In our \Mistral{} sweep, GSM8K accuracy peaks at $\beta = 0.02$ while aggregated logic-probe margins are strongly negative, illustrating that aggregate benchmarks can mask internal reasoning degradation under alignment pressure.

\subsection{R4: Hysteresis and Irreversibility}
\label{sec:results:hysteresis}

Training path affects final capability even at identical final $\beta$.
We compare two paths to $\beta = 0.01$:
\begin{itemize}
\item \textbf{Path A (Quench)}: $\beta = 0.01$ for 200 steps.
\item \textbf{Path B (Anneal)}: $\beta = 0.02$ for 200 steps, then $\beta = 0.01$ for 200 steps.
\end{itemize}

\begin{figure}[t]
\centering
\includegraphics[width=\columnwidth]{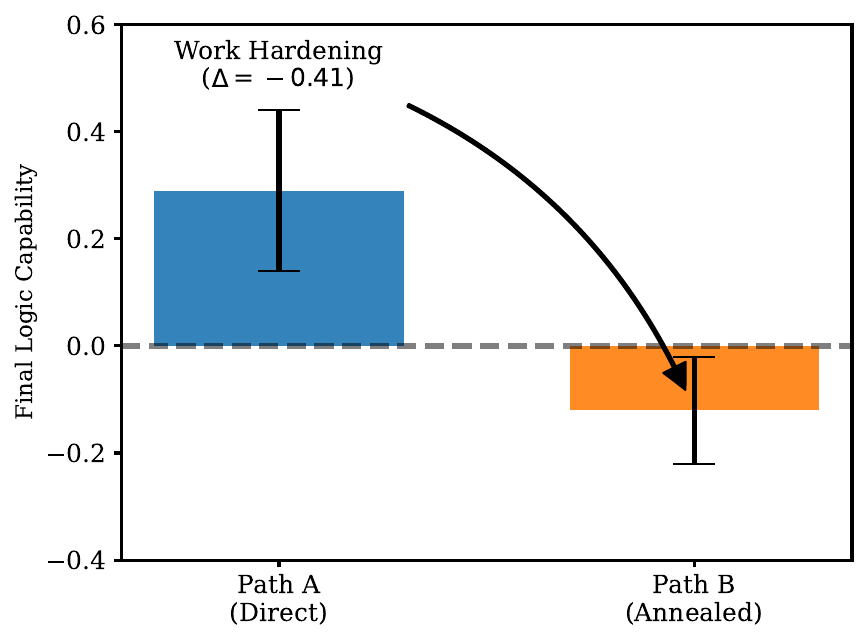}
\caption{\textbf{Hysteresis.} Path B shows significant degradation despite identical final $\beta$ (paired $d_z = 1.45$, $p = 0.032$).}
\label{fig:hysteresis}
\end{figure}

\begin{table}[t]
\caption{Hysteresis test ($n=5$). Path B yields worse capability despite identical final $\beta$.}
\label{tab:hysteresis}
\centering
\small
\begin{threeparttable}
\begin{tabular}{lcccc}
\toprule
Capability & Path A & Path B & $p$ & $d_z$ \\
\midrule
Logic & +0.29 & $-$0.12 & 0.032 & 1.45 \\
Format & $-$0.76 & $-$1.02 & 0.021 & 1.65 \\
Syco & $-$0.75 & $-$1.71 & 0.109 & 0.92 \\
\bottomrule
\end{tabular}
\begin{tablenotes}[flushleft]
\scriptsize
\item \textbf{Note:} Sycophancy does not reach significance at $\alpha=0.05$, but the effect size is large ($d_z=0.92$), suggesting possible degradation.
\end{tablenotes}
\end{threeparttable}
\end{table}

Despite \textbf{twice the training steps}, Path B yields worse capability at the same terminal $\beta$ (Figure~\ref{fig:hysteresis}, Table~\ref{tab:hysteresis}).
We report paired $t$-tests as our primary analysis (e.g., logic: $p=0.032$, $d_z=1.45$), and include a Wilcoxon signed-rank test as a robustness check (logic: $p=0.0625$).
Across all five seeds, the degradation direction is consistent for the reported capabilities, supporting genuine path dependence under our protocol.

This is analogous to \emph{work hardening} in materials: the system cannot return to its original state after traversing high-$\beta$ regions.
This persistent capability loss aligns with recent findings on the \emph{loss of plasticity} in continual learning \citep{dohare2024plasticity}, suggesting DPO can induce similar rigidity in weight space.
The implication is that ``annealing'' (starting high and cooling down) is not a safe strategy for DPO, as the high-$\beta$ trajectory induces deformation that cannot be reversed under our protocol.
\textbf{Exposure to higher-$\beta$ training can induce capability losses that persist when $\beta$ is reduced.}

\subsection{R5: Capabilities Are Bundles, Not Scalars}
\label{sec:results:probes}

Aggregate metrics obscure probe-level heterogeneity.
At $\beta = 0.01$, \Mistral{}'s logic aggregate is $+0.04$ (barely positive) but compresses dramatically different behaviors (Table~\ref{tab:probes}).

\begin{table}[t]
\caption{Probe decomposition at $\beta=0.01$. Logic aggregate (+0.04) hides 2/3 negative probes.}
\label{tab:probes}
\centering
\small
\begin{tabular}{llcc}
\toprule
Category & Probe & Score & Sign \\
\midrule
\multirow{3}{*}{Logic} & syllogism\_1 & $-$0.62 & $-$ \\
& syllogism\_2 & +2.38 & + \\
& ordering & $-$1.62 & $-$ \\
\midrule
\multirow{4}{*}{Format} & json\_simple & +0.35 & + \\
& json\_key & $-$0.04 & $-$ \\
& strict\_bool & $-$4.48 & $-$ \\
& strict\_json & +0.56 & + \\
\bottomrule
\end{tabular}
\end{table}

Within logic: \texttt{syllogism\_2} is always positive (robust); \texttt{syllogism\_1} is positive only at $\beta = 0.001$ (fragile); \texttt{ordering} is never positive (frozen).
A model ``passing'' the aggregate may fail 2/3 of constituent tasks.

Within the logic bundle itself, probes trade off: \texttt{syllogism\_1} and \texttt{syllogism\_2} show $r = -0.71$ across the $\beta$ sweep ($n=13$ grid points, $p = 0.006$).

The bundle structure extends beyond logic: sycophancy probes \texttt{flat\_earth} and \texttt{bad\_math} are anti-correlated across the $\beta$ sweep in \Mistral{} ($r=-0.50$), indicating that even within a named capability, constituent tasks can trade off under the same alignment pressure.
In \Mistral{}, these probes agree in sign \emph{only} near $\beta \approx 0.009$--$0.010$, reinforcing that this narrow band corresponds to a qualitatively distinct regime.

\subsection{R6: Structural Collapse}

\begin{figure}[t]
\centering
\includegraphics[width=\columnwidth]{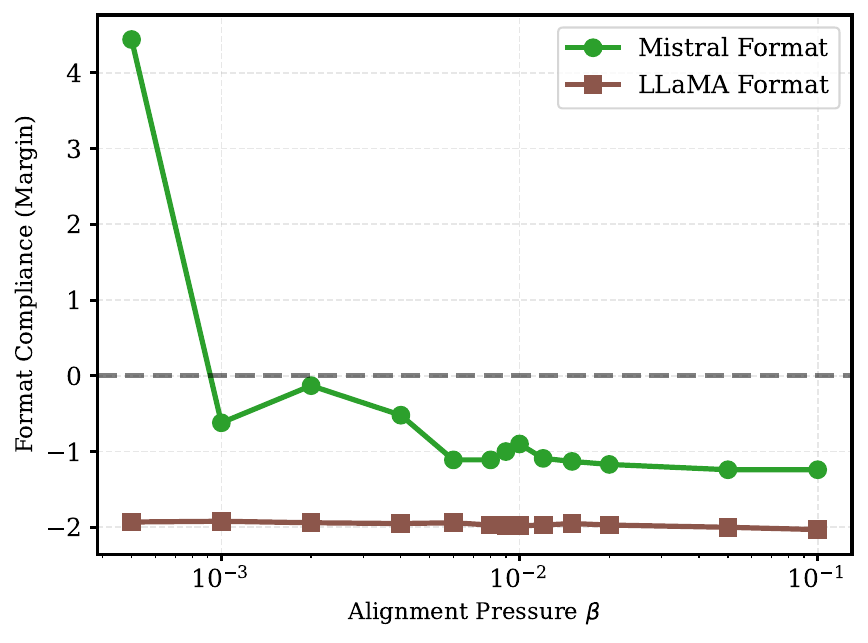}
\caption{\textbf{Structural collapse.} \Mistral{} format integrity (+4.44) degrades with $\beta$, preceding logic collapse. \Llama{} begins broken (format margin $\approx -2$ under our sweep) and never resolves.}
\label{fig:format}
\end{figure}

Format compliance probes low-level structural priors (Figure~\ref{fig:format}).
\Mistral{} begins with high integrity (+4.44 at $\beta = 0.0005$) which erodes as $\beta$ increases, preceding logic collapse.
\Llama{} begins with broken priors (format margin $\approx -2$ under our sweep) that never resolve, consistent with its selective response mode.

\subsection{R7: Co-Localized Sensitivity Without Universality}
\label{sec:results:sensitivity}

A natural concern is whether the logic-positive pocket reflects a peculiarity of \Mistral{} or a broader phenomenon.
Across three 7B families under the same sweep, only \Mistral{} exhibits sharp logic reorganization; the other architectures show anomalous but qualitatively different behavior near $\beta \approx 0.01$.

\Mistral{} reorganizes (capability sign changes), \Llama{} responds selectively (sycophancy and format reorganize while logic remains frozen), and \Qwen{} absorbs pressure smoothly (continuous trade-offs without sharp transitions).

Because the goal is to identify sensitive regions and qualitative modes rather than estimate a precise critical point, we emphasize multi-seed variance scans (\S\ref{sec:results:architecture}) and explicit path-dependence tests (\S\ref{sec:results:hysteresis}) rather than attaching error bars to phase-diagram plots.

\paragraph{Auxiliary associative scan.}
We ran an auxiliary associative-logic diagnostic on \Mistral{} and \Qwen{} (seed 1) under a faster training schedule (100 steps, $\mathrm{lr}=2\times10^{-4}$) and release the raw outputs in the reproducibility archive.
Under this probe family, \Mistral{} exhibits a sharp increase in associative margin between $\beta=0.009$ and $\beta=0.010$ ($+$175\%), co-localizing with the canonical roughness collapse, while \Qwen{} varies smoothly with small amplitude across the grid.

%%%%%%%%%%%%%%%%%%%%%%%%%%%%%%%%%%%%%%%%%%%%%%%%%%%%%%%%%%%%%%%%%%%%%%%%%%%%%%%
\section{Discussion}
%%%%%%%%%%%%%%%%%%%%%%%%%%%%%%%%%%%%%%%%%%%%%%%%%%%%%%%%%%%%%%%%%%%%%%%%%%%%%%%

\subsection{Scope and Boundaries}

We do not claim universality classes, critical exponents, or mechanisms. 
We claim only what our controlled experiments support: at 7B scale and under a fixed DPO recipe, varying $\beta$ can produce sharp discontinuities, seed-sensitive boundaries, and path-dependent outcomes.
The physics vocabulary is an organizing framework, not a thermodynamic assertion.

\paragraph{What our comparisons do and do not imply.}
This paper is a controlled study of how DPO alignment pressure reconfigures \emph{capability probes} under a fixed recipe at 7B scale, with emphasis on \emph{logic}-probe dynamics in \Mistral{}.
Our cross-architecture comparisons do \emph{not} imply that \Llama{} or \Qwen{} are globally unresponsive to alignment; rather, under the same sweep and within our probe set, we do not observe a \Mistral{}-like \emph{logic-positive pocket} (sharp, transient logic reorganization co-localized with roughness discontinuities).
Both models exhibit changes in non-logic probe categories across $\beta$ (Table~\ref{tab:three_arch}), suggesting architecture- and capability-specific response modes that warrant targeted follow-up.

\paragraph{Why ``phase transitions'' is appropriate.}
Operationally, we use ``phase transition'' to denote: (i) a qualitative change in a summary statistic (here, probe margins) over a narrow control-parameter interval, (ii) path dependence (hysteresis), and (iii) high susceptibility near the boundary (seed-sensitive outcomes).
We observe all three in \Mistral{}: logic sign change over narrow $\beta$; path-dependent final states; opposite outcomes at fixed $\beta$ across seeds.
This operational usage is consistent with prior work on grokking \citep{power2022grokking} and sudden capability transitions \citep{chen2024sudden}.

\subsection{Three Response Modes}

Based on the observed variance and stability profiles, we classify the architectures into three response modes:

\paragraph{Plastic (\Mistral{}).}
Capable of reorganization, allowing for logic-positive pockets but risking collapse. High variance near transitions. Outcomes are seed-sensitive.

\paragraph{Selective (\Llama{}).}
Capability-specific rigidity: logic remains frozen across the $\beta$ sweep, but sycophancy and format probes reorganize in the same critical region as \Mistral{}.
The learning rate test confirms logic-specific resistance; other capabilities remain plastic.
This suggests architectural compartmentalization of alignment response.

\paragraph{Smooth (\Qwen{}).}
Capabilities trade off gradually across the $\beta$ sweep without sharp discontinuities in our probe margins, consistent with a comparatively smooth response mode under this training recipe.

\noindent This framework explains why a single ``best $\beta$'' does not exist: the optimal pressure depends on the model's inherent alignment viscosity.

\subsection{Mechanistic Hypotheses}

\paragraph{Capacity displacement.}
Surface capabilities emerge at lower $\beta$ than reasoning. Under high pressure, the model reallocates capacity to strict syntax and tone constraints at the expense of fragile reasoning circuits. This is consistent with surface behaviors (e.g., sycophancy/negation) shifting at lower $\beta$ than logic, while structural format integrity can erode as $\beta$ increases.

\paragraph{Manifold collapse.}
Strong negative margin-capability correlation suggests DPO forces the model onto a ``preference manifold'' geometrically misaligned with the ``truth manifold.'' 
As $\beta$ increases, the model collapses onto a lower-dimensional subspace, losing features required for compositional reasoning.
This connects to findings that RLHF can reduce output diversity \citep{kirk2024rlhf}, though we observe the effect at the level of internal capability rather than surface generation.

\paragraph{KL-gradient comparability.}
One plausible hypothesis is that the sensitive band near $\beta \approx 0.01$ marks a regime where the KL-related term becomes comparable in magnitude to the preference term in the update geometry.
Below this band, the KL-related effect may be weak; above it, it may dominate and constrain updates.
If so, the transition reflects a qualitative change in trajectory rather than a smooth trade-off.
Future work can test this hypothesis by directly tracking per-term gradient norms and angles during training.

\subsection{Safety Implications}

\begin{enumerate}
\item \textbf{Margin is not a safety proxy.} If margin anticorrelates with capability, standard early-stopping may select worse models.
\item \textbf{Benchmarks can mask degradation.} Surface metrics improve while reasoning degrades; systems appear aligned but fail on integration.
\item \textbf{Single-seed evaluation is insufficient.} With 2--3 orders of magnitude variance difference, single runs provide no distributional guarantee.
\item \textbf{Training order matters.} Hysteresis means annealing $\neq$ quenching. Over-alignment induces losses that do not recover under our protocol.
\item \textbf{Capabilities are bundles.} A model can pass aggregates while failing most constituent probes.
\end{enumerate}

\begin{table}[t]
\centering
\small
\caption{Practical checklist for alignment hyperparameter sweeps.}
\label{tab:checklist}
\begin{tabular}{@{}p{0.92\columnwidth}@{}}
\toprule
\textbf{Recommendations for practitioners} \\
\midrule
1. Do not select $\beta$ using margin alone; it may anticorrelate with capability. \\[2pt]
2. Evaluate capabilities at multiple abstraction levels (e.g., format \emph{and} logic). \\[2pt]
3. Run $\geq 3$ seeds near any performance peak; single runs are unreliable near sensitive $\beta$. \\[2pt]
4. Avoid annealing through high-$\beta$ regions without rollback validation. \\[2pt]
5. Treat benchmark gains near any empirically identified sensitive $\beta$ band (e.g., $\beta \approx 0.01$ in our 7B sweeps) as non-diagnostic without capability-probe validation. \\
\bottomrule
\end{tabular}
\end{table}

\subsection{Limitations}

\begin{itemize}
\item \textbf{Scale}: We study 7B models because they permit dense scanning of the $\beta$ landscape with tractable compute. Prior work shows that capability transitions in neural networks become increasingly sharp with scale, exhibiting step-like onsets and narrowed transition regions \citep{wei2022emergent,caballero2022discrete}, potentially making ordered pockets narrower and harder to target. Our 7B results may underestimate the difficulty of alignment tuning at frontier scale. At larger scales, optimization operates under tighter compute--data tradeoffs \citep{hoffmann2022training}, potentially increasing sensitivity to alignment hyperparameters.
\item \textbf{Resolution}: Critical boundaries require finer sweeps than our 13-point grid; we increased resolution only near $\beta \approx 0.006$--$0.015$.
\item \textbf{Coverage}: 14 probes across 5 categories is diagnostic, not comprehensive.
\item \textbf{Mechanism}: We provide empirical constraints, not internal explanations.
\item \textbf{Generalization}: We observe co-localized sensitivity at $\beta \approx 0.01$ for these three 7B models; whether this threshold generalizes to other scales or training regimes remains open.
\end{itemize}

%%%%%%%%%%%%%%%%%%%%%%%%%%%%%%%%%%%%%%%%%%%%%%%%%%%%%%%%%%%%%%%%%%%%%%%%%%%%%%%
\section{Related Work}
%%%%%%%%%%%%%%%%%%%%%%%%%%%%%%%%%%%%%%%%%%%%%%%%%%%%%%%%%%%%%%%%%%%%%%%%%%%%%%%

\paragraph{DPO and variants.}
DPO \citep{rafailov2023direct} reformulates RLHF as supervised learning. Variants include IPO \citep{azar2024general}, KTO \citep{ethayarajh2024kto}, ORPO \citep{hong2024orpo}, and SimPO \citep{meng2024simpo}. Recent work documents instabilities including length bias \citep{park2024disentangling,xu2024contrastive}. None systematically study $\beta$ as a control parameter inducing capability transitions.

\paragraph{Emergence.}
\citet{wei2022emergent} document emergent abilities versus scale; \citet{schaeffer2023emergent} argue these may be metric artifacts. We show capability changes at \emph{fixed scale}, a distinct phenomenon given observed non-monotonicity and hysteresis.

\paragraph{Phase transitions in ML.}
Sharp transitions have been identified in double descent \citep{belkin2019reconciling}, grokking \citep{power2022grokking}, scaling laws \citep{hoffmann2022training}, and sudden capability emergence \citep{chen2024sudden}. \citet{choromanska2015loss} connected deep network loss surfaces to spin-glass landscapes. We extend this perspective to post-training alignment, treating $\beta$ as an external control parameter analogous to temperature.

\paragraph{Goodhart, forgetting, and plasticity.}
Proxy-objective decoupling is a known optimization risk \citep{goodhart1984problems}. \citet{gao2023scaling} characterize reward overoptimization scaling laws; \citet{perez2022discovering} established that RLHF can induce sycophancy.
Prior work documents capability degradation as an alignment cost \citep{askell2021general,lin2024alignment}, and \citet{kirk2024rlhf} show RLHF can reduce output diversity.
\citet{dohare2024plasticity} demonstrate that networks can lose the ability to learn after certain training regimes, a phenomenon we observe manifesting as hysteresis.
Our results extend this literature: the alignment tax is \emph{non-monotonic} and can be \emph{persistent under our protocol}, connecting to catastrophic forgetting \citep{kirkpatrick2017overcoming} but with sharp transitions rather than gradual decay.

%%%%%%%%%%%%%%%%%%%%%%%%%%%%%%%%%%%%%%%%%%%%%%%%%%%%%%%%%%%%%%%%%%%%%%%%%%%%%%%
\section{Conclusion}
%%%%%%%%%%%%%%%%%%%%%%%%%%%%%%%%%%%%%%%%%%%%%%%%%%%%%%%%%%%%%%%%%%%%%%%%%%%%%%%

Alignment under DPO is not a smooth curve but a landscape with ridges, valleys, and cliffs.
At fixed 7B scale, varying $\beta$ induces capability emergence, collapse, and path-dependent degradation in \Mistral{}, including a narrow logic-positive pocket in which logic becomes transiently positive despite a simultaneous decrease in the preference margin.

Under our fixed optimization settings, we observe a co-localized region near $\beta \approx 0.01$ where all three architectures exhibit noticeable changes in at least some probes, but with architecture-specific expression: \Mistral{} reorganizes (capability sign changes), \Llama{} responds selectively (sycophancy and format shift while logic remains frozen), and \Qwen{} absorbs pressure smoothly (continuous trade-offs without sharp transitions).
The stress point is shared; the response mode is not.

Perhaps most concerning, benchmark performance can improve even as internal reasoning degrades, meaning margin-based early stopping, single-seed runs, and aggregate metrics can select capability-impaired models that appear well-aligned.

We propose a \emph{phase-diagram methodology}: (1)~map capability-vs-$\beta$ structure before deployment; (2)~validate with multiple seeds near sensitive regions; (3)~monitor capabilities, not just margins or benchmarks, during training; (4)~avoid curricula crossing collapse boundaries.

The practical question is not ``what $\beta$ maximizes margin?'' but ``can we certify the model remains capability-ordered under realistic training paths?''

Future work should test whether the sensitivity threshold shifts predictably with model scale, and whether logic-positive pockets can be deliberately stabilized through curriculum design.

\section*{Broader Impact}

This work advances the scientific understanding of alignment dynamics in large language models by empirically characterizing non-monotonic and path-dependent effects of Direct Preference Optimization.
Its primary contribution is methodological: providing tools and diagnostics for more reliable evaluation of alignment hyperparameters.

The findings have potential positive societal impact by informing safer deployment practices, highlighting failure modes where standard alignment metrics may be misleading, and encouraging more robust evaluation prior to real-world use.
The work does not introduce new model architectures, datasets, or deployment mechanisms, and does not directly enable harmful applications beyond those already associated with large language models.

As with all research on model alignment, there is a dual-use consideration: insights into optimization-induced failures could be misused to target brittle regions of hyperparameter space and induce failure modes.
However, we believe transparency about such dynamics is essential for improving robustness and safety, and that the benefits of identifying and mitigating these risks outweigh potential misuse.

%%%%%%%%%%%%%%%%%%%%%%%%%%%%%%%%%%%%%%%%%%%%%%%%%%%%%%%%%%%%%%%%%%%%%%%%%%%%%%%
\section*{Reproducibility Statement}
%%%%%%%%%%%%%%%%%%%%%%%%%%%%%%%%%%%%%%%%%%%%%%%%%%%%%%%%%%%%%%%%%%%%%%%%%%%%%%%

\textbf{Models.}
All experiments use publicly available 7B-scale checkpoints:
\Mistral{} (\href{https://huggingface.co/mistralai/Mistral-7B-v0.1}{\nolinkurl{mistralai/Mistral-7B-v0.1}}),
\Llama{} (\href{https://huggingface.co/NousResearch/Llama-2-7b-hf}{\nolinkurl{NousResearch/Llama-2-7b-hf}}),
and \Qwen{} (\href{https://huggingface.co/Qwen/Qwen1.5-7B}{\nolinkurl{Qwen/Qwen1.5-7B}}).

\textbf{Protocols.} We run a \emph{canonical} $\beta$ sweep where each $\beta$ point is trained from a fresh base checkpoint under an identical DPO recipe. Hysteresis experiments intentionally violate the fresh-start principle and are labeled as Path A/B. An auxiliary associative scan uses a faster training schedule (100 steps, $\mathrm{lr}=2\times10^{-4}$) and is released separately.

\textbf{Training hyperparameters.} Unless otherwise stated: learning rate $5\times 10^{-5}$, batch size 4, 200 optimizer steps, AdamW, LoRA rank 8 (alpha 16, dropout 0.05). Hysteresis uses 400 total steps (200 + 200).
Where multi-seed experiments are reported, we use an explicit seed list (seeds 1--5); otherwise single-seed sweeps use seed 1.
 All $\beta$ grids and seed lists used to generate each figure/table are provided in the associated repository.

\textbf{Evaluation.} Capability is measured by length-normalized log-probability margins on fixed probe sets (logic, arithmetic, format, sycophancy, negation). Positive margin indicates correct behavior under the probe scoring rule. Plots and tables are generated from the raw JSON logs produced by the evaluation scripts. For GSM8K, we observed a degenerate zero score at $\beta=0.002$ under our evaluation pipeline; we treat this point as missing data in benchmark--probe correlation analyses.

\textbf{Artifacts.} We provide an anonymized repository containing the experimental protocol (including all hyperparameter settings), runnable experiment scripts, environment requirements, raw per-run JSON outputs, and plotting code sufficient to reproduce every figure and table in this paper:
\href{https://github.com/anon-repo-317/experiments}{https://github.com/anon-repo-317/experiments}.

%%%%%%%%%%%%%%%%%%%%%%%%%%%%%%%%%%%%%%%%%%%%%%%%%%%%%%%%%%%%%%%%%%%%%%%%%%%%%%%
\bibliography{references}
\bibliographystyle{icml2026}
\end{document}